%% file: acl_latex.tex
\pdfoutput=1
\documentclass[11pt]{article}

\usepackage[final]{acl}

\usepackage{times}
\usepackage{latexsym}

\usepackage[T1]{fontenc}
\usepackage[utf8]{inputenc}

\usepackage{microtype}

\usepackage{inconsolata}

\usepackage{graphicx}
\usepackage{fontawesome5}
\usepackage{lipsum}
\usepackage{amssymb}
\usepackage{multirow}
\usepackage{graphicx}
\usepackage{amsmath}
\usepackage{tabularx}
\usepackage{booktabs}

\newcommand{\titlelogo}{\raisebox{-0.9ex}{\includegraphics[height=24pt]{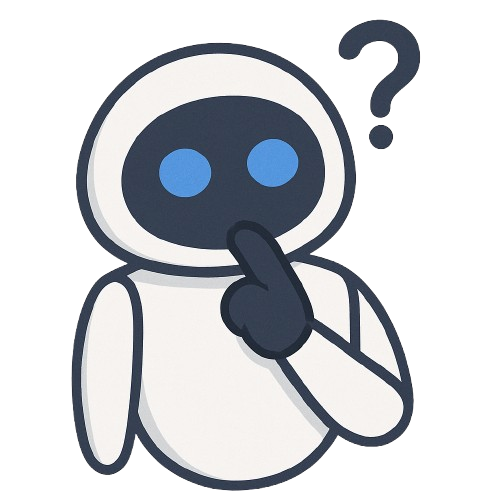}}}

\title{\titlelogo\hspace{-0.2em} QEVA: A Reference-Free Evaluation Metric
for Narrative Video Summarization with Multimodal Question Answering}

\author{Woojun Jung \and Junyeong Kim\thanks{Corresponding author.}\\
  Department of Artificial Intelligence, Chung-Ang University\\
  \texttt{\{svvma91, junyeongkim\}@cau.ac.kr}}

\begin{document}
\maketitle
\input{content/0_abstract}
\input{content/1_intro}
\input{content/2_related}
\input{content/3_method_1}
\input{content/4_exp}

\input{content/5_con}
\input{content/6_lim}
% Bibliography entries for the entire Anthology, followed by custom entries
%\bibliography{anthology,custom}
% Custom bibliography entries only
\bibliography{custom}

\appendix
\input{content/appendix}

\end{document}

%% file: content/0_abstract.tex
\begin{abstract}
% video-to-text summarization 보다 더 좋은 표현이 없을까나

Video-to-text summarization remains underexplored in terms of comprehensive evaluation methods. Traditional n-gram overlap-based metrics and recent large language model (LLM)-based approaches depend heavily on human-written reference summaries, limiting their practicality and sensitivity to nuanced semantic aspects. In this paper, we propose QEVA, a reference-free metric evaluating candidate summaries directly against source videos through multimodal question answering. QEVA assesses summaries along three clear dimensions: Coverage, Factuality, and Chronology. We also introduce MLVU(VS)-Eval, a new annotated benchmark derived from the MLVU dataset, comprising 800 summaries generated from 200 videos using state-of-the-art video-language multimodal models. This dataset establishes a transparent and consistent framework for evaluation. Experimental results demonstrate that QEVA shows higher correlation with human judgments compared to existing approaches, as measured by Kendall’s $\tau_b$, $\tau_c$, and Spearman’s $\rho$. We hope that our benchmark and metric will facilitate meaningful progress in video-to-text summarization research and provide valuable insights for the development of future evaluation methods.\footnote{Our evaluation metric (QEVA) is available at \url{https://github.com/jungnerd/QEVA}}

\end{abstract}

%% file: content/1_intro.tex
\section{Introduction}
\label{sec:introduction}
\input{attachments/figure1.tex}

Text-based video summarization has become increasingly crucial due to the explosive growth of video content and significant advances in Video-Large Multimodal Models (Video-LMMs). Despite substantial progress in generating comprehensive textual summaries from videos, insufficient attention has been paid to reliably evaluating the quality of these summaries. Currently, evaluation primarily relies on reference-based metrics such as ROUGE or METEOR, which compare generated summaries to reference texts produced by human annotators. However, acquiring accurate and detailed reference summaries for videos requires considerable human effort and resources, making the process both costly and inefficient at scale. Consequently, the absence of efficient evaluation approaches has become a major obstacle to the advancement of video summarization research.  

\input{attachments/figure2}

While reference-free metrics have gained considerable attention in text summarization tasks, these methods typically compare the generated summaries directly with the source text, leveraging linguistic similarities or embeddings. Unfortunately, extending this concept to videos is inherently challenging due to the fundamental modality gap - videos are spatio-temporal and multimodal, and thus cannot be directly compared to textual summaries in a straightforward manner. Moreover, recent attempts at multimodal evaluation, such as CLIPScore or LLM-based comparisons used in benchmarks like MLVU, still rely heavily on reference summaries or fail to reliably capture important aspects such as chronological fidelity or factual accuracy. Thus, there is a pressing need for a novel, effective, and fully reference-free evaluation paradigm tailored specifically for video summarization.  

To address these challenges, we propose a \textbf{Q}uestion-answering based \textbf{E}valuation metric for \textbf{V}ideo summ\textbf{A}rization, QEVA. QEVA is built upon the intuitive principle that a high-quality video summary should be able to substitute the original video content effectively. Based on this principle, we identify three critical dimensions for evaluating summaries: Coverage (capturing all essential content), Chronology (chronological fidelity; preserving the order of events), and Factuality (ensuring factual correctness). QEVA leverages Video-LMMs to automatically generate relevant questions from the original video content across these dimensions and employs LLMs to answer these questions using only the generated summaries. By measuring the accuracy of these answers, QEVA quantitatively assesses summary quality without the need for any human reference summaries.

Furthermore, we present MLVU(VS)-Eval, a new annotated evaluation dataset derived from the MLVU benchmark, containing 800 summaries generated by Video-LMMs such as GPT-4o, QwenVL, InternVL, and Video-LLaVA. Each summary has been rigorously annotated by multiple human evaluators according to Coverage, Chronology, and Factuality, demonstrating strong inter-annotator agreement (Krippendorff's $\alpha$ = 0.68). Our comprehensive experiments show that QEVA consistently exhibits the highest correlation with human judgments compared to existing reference-based and multimodal evaluation methods (e.g., ROUGE, METEOR, BERTscore).

By introducing QEVA and the MLVU(VS)-Eval dataset, this work pioneers the first systematic exploration of reference-free evaluation for video-to-text summarization. Our approach not only addresses the pressing scalability and practicality issues but also sets a clear foundation for future advancements in multimodal summarization research, facilitating rapid and reliable assessment of Video-LMMs in real-world applications, including content platforms, news summarization, and automated content generation services.

%% file: attachments/figure1.tex
\begin{figure}[t!]
\centering
    % \includegraphics[width=\linewidth]{attachments/figure1.pdf} % 이미지 로드 주석 처리
    % \framebox[.95\linewidth]{\rule{0pt}{10cm}} % 너비는 \linewidth, 높이는 5cm인 빈 박스 (높이는 조절 가능)
    \includegraphics[width=\linewidth]{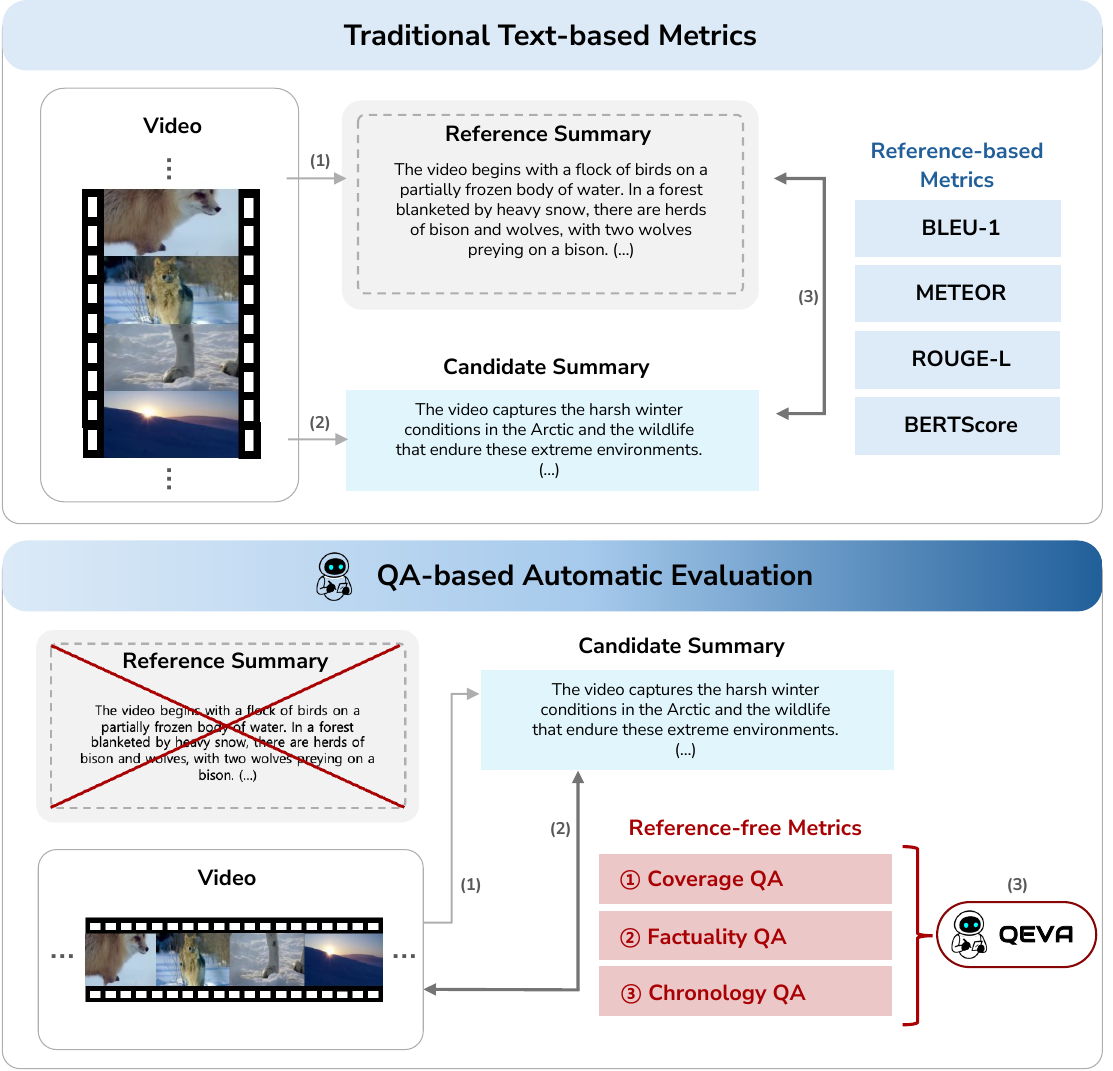}
    \caption{Overview of existing video summarization evaluation approaches and our proposed method, QEVA.
(Top) Traditional reference-based metrics rely solely on text-to-text comparisons between candidate and human-written reference summaries, often failing to capture nuanced semantic and multimodal content. (Bottom) QEVA leverages a fully reference-free multimodal question-answering pipeline (Coverage QA, Factuality QA, Chronology QA) to directly evaluate candidate summaries against source videos, enabling a more comprehensive and semantically grounded assessment.
\textbf{Takeaway:} QEVA provides a more accurate and scalable alternative by eliminating the reliance on human-written reference summaries and directly assessing summaries against source video content.}
    \label{fig:vtsscore_overview}
  \end{figure}

%% file: attachments/figure2.tex
\begin{figure*}[t]
    \centering
    \includegraphics[width=\textwidth]{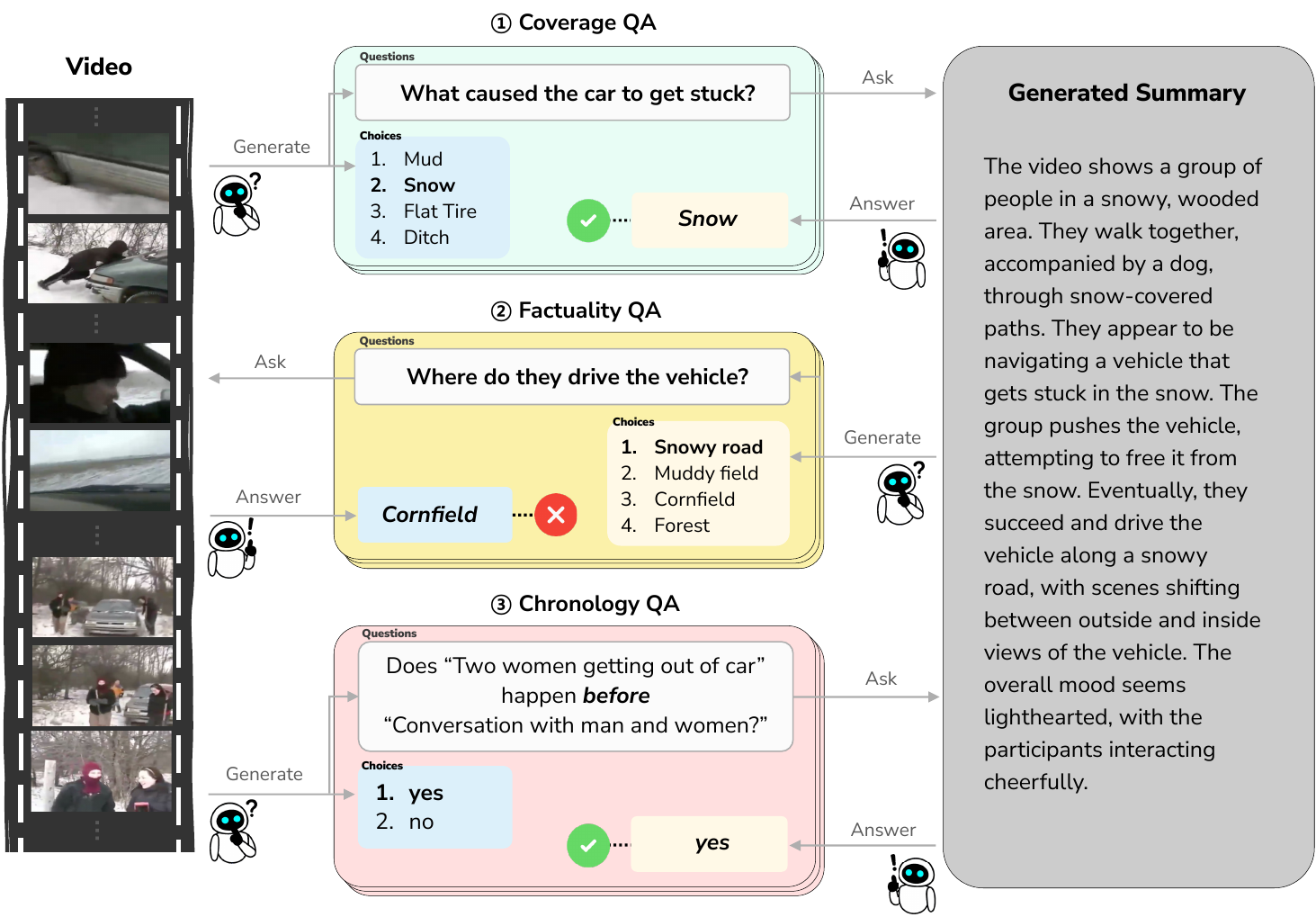}
    \caption{Detailed illustration of QEVA's multimodal question-answering methodology. Given a video and candidate summary, QEVA evaluates the summary across three distinct dimensions: (1) Coverage (whether the summary comprehensively covers key video content), (2) Factuality (accuracy of the information presented), and (3) Chronology (i.e.,chronological fidelity; correctness of event ordering). QEVA generates tailored QA pairs through a structured pipeline involving question generation, question answering, and answer correctness checking.
\textbf{Takeaway:} QEVA explicitly decomposes video summary quality into three complementary dimensions, enabling nuanced and interpretable evaluation.}
    \label{fig:qeva_pipeline}
    \end{figure*}

%% file: content/2_related.tex
\section{Related Work}

\subsection{Automated Evaluation of Summarization}

Automated evaluation has traditionally relied on lexical overlap metrics such as BLEU~\cite{papineni2002bleu}, ROUGE~\cite{lin2004rouge}, METEOR~\cite{banerjee2005meteor}, CIDEr~\cite{vedantam2015cider}, and SPICE~\cite{anderson2016spice}. While computationally efficient and widely used, these reference-based metrics primarily measure surface-level similarity between candidate and human-written reference texts, often failing to capture deeper semantic consistency, factual correctness, or coherence~\cite{kryscinski2019evaluating, kasai2022transparenthumanevaluationimage}. 

To address these limitations, embedding-based metrics such as BERTScore~\cite{zhang2020bertscore} and CLIPScore~\cite{hessel2022clipscorereferencefreeevaluationmetric} have been proposed to measure semantic similarity using learned embeddings. However, these methods still require reference summaries, limiting their practical applicability and scalability. More recently, Large Language Models (LLMs) have been leveraged as evaluators ("LLM-as-a-Judge"), demonstrating promising correlation with human judgments~\cite{zheng2023judgingllmasajudgemtbenchchatbot}, but they introduce challenges such as biases, reproducibility issues, and high API costs~\cite{fu2023gptscore}.

Our proposed QEVA metric addresses these challenges by employing a fully reference-free evaluation approach, directly comparing generated summaries against source videos through multimodal question-answering, thus providing a more semantically grounded and scalable alternative.

\subsection{QA-based Evaluation Metrics}

Question-answering (QA) based evaluation methods assess the semantic quality of generated texts by testing their ability to answer questions derived from relevant contexts~\cite{wang2020askingansweringquestionsevaluate}. Unlike traditional lexical metrics, QA-based evaluation directly probes textual outputs for factuality, coverage, and relevance. Examples include FEQA~\cite{Durmus_2020} and QAFactEval~\cite{fabbri2022qafacteval}, which evaluate factual consistency by comparing answers derived from summaries and source documents.

Reference-free QA metrics such as QuestEval~\cite{scialom2021questeval} and $Q^2$~\cite{honovich2021q2evaluatingfactualconsistency} offer increased scalability by eliminating the dependency on human-written references. QuestEval integrates precision-oriented (summary-based) and recall-oriented (source-based) QA components, achieving strong correlation with human judgments. Recently, the TIFA metric~\cite{hu2023tifaaccurateinterpretabletexttoimage} extended QA-based evaluation to text-to-image synthesis by generating questions from textual prompts and assessing image fidelity through visual QA.

Our work builds upon these advances by introducing QEVA, which uniquely applies multimodal QA to narrative video summarization, explicitly measuring Coverage, Factuality, and Chronology, thereby providing a comprehensive and reference-free evaluation framework tailored specifically for multimodal summarization.

\input{attachments/figure3}

%% file: attachments/figure3.tex
\begin{figure*}[t]
\centering
\includegraphics[width=\textwidth]{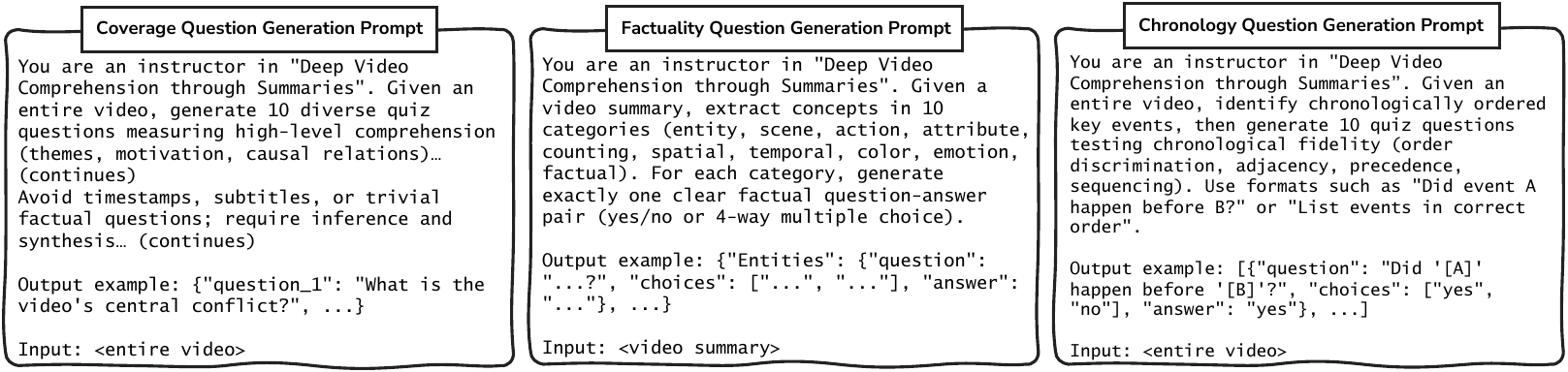}
\caption{Detailed prompts used by QEVA for multimodal question-answer generation across three distinct evaluation dimensions: (a) Coverage, (b) Factuality, and (c) Chronology.
(a) Coverage QA prompts instruct a Video-LMM to generate high-level, inference-driven questions that assess whether the summary comprehensively captures essential content, themes, and events from the source video.
(b) Factuality QA prompts guide an LLM to first extract critical factual elements explicitly mentioned in the summary (such as entities, actions, attributes), and subsequently generate targeted questions designed to verify factual accuracy.
(c) Chronology QA prompts outline a structured workflow for event segmentation from the source video, sampling adjacent and distant pairs of events, and generating questions to evaluate whether the summary preserves the correct chronological order of these events.
\textbf{Takeaway:} QEVA employs carefully structured prompts tailored to each evaluation dimension, enabling a systematic and comprehensive assessment of video summary quality along the axes of Coverage, Factuality, and Chronology.}
\label{fig:qa_prompts}
\end{figure*}

%% file: content/3_method_1.tex
\section{The QEVA Method}
\label{sec:method}

We introduce \textbf{QEVA}, a novel evaluation framework designed to assess how comprehensively and faithfully a textual video summary captures the original video's content. QEVA is grounded in the principle that \textit{"a good summary should serve as an effective substitute for the source video"}. To operationalize this intuition, QEVA employs multimodal question answering (QA) as its core mechanism: a high-quality summary should allow accurate answering of critical questions derived from the source video content. Unlike prior work relying solely on direct LLM-based judgments, QEVA integrates both Video-LMMs and LLMs within a collaborative pipeline, thus mitigating the inherent limitations of Video-LMMs as direct evaluators (see Section~\ref{sec:ablation} for detailed analyses). Given a source video and candidate summary, QEVA outputs a multidimensional quality score reflecting the summary's \textit{coverage}, \textit{factuality}, and \textit{chronology (chronological fidelity)}. Figure~\ref{fig:qeva_pipeline} provides an overview of our QEVA workflow.

\subsection{Evaluation Criteria}
Inspired by prior work in text summarization evaluation (e.g., SummEval) and recent multimodal captioning evaluation metrics (e.g., ACCR in G-VEval), we define three complementary criteria to comprehensively evaluate video summaries: \textbf{Coverage}, \textbf{Factuality}, and \textbf{Chronology}.

\paragraph{Coverage.}
Coverage measures whether the summary includes essential content, key events, and the main messages of the source video. A high-coverage summary should comprehensively represent all core information and omit no major events or salient points.

\paragraph{Factuality.}
Factuality assesses whether the summary accurately reflects details present in the source video without introducing hallucinated or unsupported information. Good factuality implies all claims in the summary can be directly verified from the video.

\paragraph{Chronology.}
Chronology (chronological fidelity) evaluates whether a summary preserves the source video's order of events and correctly interprets non-linear structures (e.g., flashbacks/flash-forwards). 
This criterion concerns ordering only; fine-grained timing such as duration or pace is not evaluated.

These criteria are mutually complementary. A summary may comprehensively cover all major events (\textit{Coverage}) but misrepresent critical details (\textit{Factuality}), or accurately summarize events but distort their order (\textit{Chronology}). By explicitly defining these distinct evaluation dimensions, QEVA ensures consistent and interpretable evaluation, minimizing subjective interpretation and inter-annotator variance.

\subsection{Question-Answer Generation}
\label{sec:qa-generation}
For each evaluation criterion, QEVA generates tailored question-answer (QA) pairs to probe summary quality. Figure~\ref{fig:qa_prompts} illustrates example prompts for QA generation, with complete prompts provided in Appendix~\ref{sec:full-prompts}.

\paragraph{Coverage QA Generation.}
We prompt a Video-LMM with the full source video and a specialized instruction to generate $N$ diverse, high-level questions. These prompts encourage synthesis and inference-based questions that target main events, causal relations, and overarching themes, explicitly avoiding superficial factual or timestamp-specific queries.

\paragraph{Factuality QA Generation.}
We prompt an LLM to extract salient elements from the candidate summary and categorize them (entities, actions, attributes, counting, etc.). Then, we generate targeted factual queries that require answers strictly supported by the summary itself, enabling verification of summary correctness against the source video.

\paragraph{Chronology QA Generation.}
We first employ a Video-LMM to extract a chronological sequence of key events from the video. Subsequently, we sample pairs of events (adjacent and non-adjacent) and generate three types of chronological questions: (1) order verification (yes/no), (2) temporal precedence (multiple-choice), and (3) event sequence sorting tasks. These questions explicitly evaluate if the summary accurately preserves event ordering.

\subsection{Question and Answer Filtering}
To ensure the quality and discriminative power of generated QA pairs, we introduce a two-stage automatic filtering step. This process is designed to eliminate questions that either do not require the source context (i.e., are trivial) or are inherently flawed (i.e., ambiguous or low-quality). Specifically, we employ an alternative Video-LMM or LLM (distinct from the model used for generating the QA pairs) to answer each question under two conditions:

\begin{itemize}
    \item \textbf{Trivial Filtering:} If the alternative model correctly answers a question without any context (using only the question and answer choices), we consider it trivial and remove it. This step is vital to ensure that QEVA genuinely tests for faithfulness to the provided context rather than the model's own parametric knowledge.
    \item \textbf{Low-quality Filtering:} If the model fails to answer correctly even when provided with the appropriate context (the source video or summary), the question is considered ambiguous or unanswerable and is thus discarded. This removes noise from the evaluation.
\end{itemize}

This two-stage filtering ensures retained QA pairs are neither trivial nor excessively ambiguous.

\subsection{QEVA Score Computation}
For each criterion, QEVA computes an evaluation score as the proportion of correctly answered questions from the filtered set. Let \(Q_{Cov}\), \(Q_{Fact}\), and \(Q_{Chrono}\) denote the filtered QA sets for Coverage, Factuality, and Chronology, respectively.

\begin{itemize}
    \item For \textbf{Coverage} and \textbf{Chronology}, the summary \(S\) is used as the context. The scores are the proportions of correctly answered questions:
    \begin{align}
        \mathrm{Score}_{Cov}(S, Q_{Cov}) &= \frac{\left|Q_{Cov, \text{correct}}\right|}{\left|Q_{Cov}\right|} \\
        \mathrm{Score}_{Chrono}(S, Q_{Chrono}) &= \frac{\left|Q_{Chrono, \text{correct}}\right|}{\left|Q_{Chrono}\right|}
    \end{align}
    \item For \textbf{Factuality}, the source video \(V\) is used as the context. The score is the proportion of questions confirmed as accurate:
    \begin{equation}
        \mathrm{Score}_{Fact}(V, Q_{Fact}) = \frac{\left|Q_{Fact, \text{correct}}\right|}{\left|Q_{Fact}\right|}
    \end{equation}
\end{itemize}

The final QEVA score is the arithmetic mean of these three component scores. This provides a final score on a normalized 0-to-1 scale. Formally, for a summary \(S\) and video \(V\), the score is defined as:
\begin{equation}
\begin{split}
\mathrm{QEVA}(S, V) = \bigl( & \mathrm{Score}_{Cov}(S, Q_{Cov}) \\
& + \mathrm{Score}_{Fact}(V, Q_{Fact}) \\
& + \mathrm{Score}_{Chrono}(S, Q_{Chrono}) \bigr) / 3
\end{split}
\end{equation}

\subsection{Implementation Details}
For our default experimental setup, we use Gemini-1.5 Pro as the primary Video-LMM and GPT-4o as the primary LLM for QA generation and answering. Alternative settings with open-source models (e.g., Qwen2.5-VL, InternVL3, LLaMA-3.1, Gemma-3) are also explored in Section~\ref{sec:ablation}. All prompts, hyperparameters, and scripts necessary for full reproducibility are included in the supplementary material and will be publicly released upon acceptance.

%% file: content/4_exp.tex
\input{attachments/table1}
\input{attachments/table2}

\section{Experiments}
\label{sec:experiments}

We conduct comprehensive experiments to validate the effectiveness of our proposed QEVA metric. We first introduce \textbf{MLVU(VS)-Eval}, a novel benchmark dataset for evaluating video-to-text summarization metrics (\S\ref{sec:dataset}). We then compare QEVA against existing metrics in terms of correlation with human judgments (\S\ref{sec:correlation}). Additionally, we analyze model-wise performance of QEVA (\S\ref{sec:modelwise-corr}), and demonstrate the robustness of QEVA via ablation studies (\S\ref{sec:ablation}).

\subsection{MLVU(VS)-Eval: A Novel Benchmark for Evaluating Video Summarization Metrics} \label{sec:dataset}

Existing evaluation datasets for video-to-text summarization lack human annotations, limiting accurate metric evaluation. To address this, we propose \textbf{MLVU(VS)-Eval}, a novel human-annotated dataset built upon the MLVU benchmark~\citep{zhou2024mlvu}.

We select 200 video clips (average length $\sim$15 minutes) from the MLVU Video Summarization task. Each video has a human-written reference summary. We generate candidate summaries using four widely-used Video-LMM models: GPT-4o, InternVL3-8B, Qwen2.5-VL-7B, and Video-LLaVA-7B, resulting in 800 candidate summaries. For our evaluation, we recruited 20 annotators (comprising graduate and undergraduate students) to assess each summary based on three criteria: \textit{Coverage}, \textit{Factuality}, and \textit{Chronology}, using a 5-point Likert scale. Each summary received evaluations from two independent annotators. The inter-annotator agreement, measured using Krippendorff's $\alpha$, is 0.68, indicating substantial reliability of the annotations.

\subsection{Correlation with Human Judgments} \label{sec:correlation}

We compare QEVA with various existing evaluation metrics in terms of correlation with human judgments.

\paragraph{Baseline Metrics.} 
We compare QEVA with several representative categories of evaluation metrics: reference-based n-gram overlap metrics, embedding-based similarity metrics, and LLM-based evaluation approaches. For the n-gram and embedding-based metrics, we report results using widely adopted methods in each category. Additionally, we introduce two novel baselines: (1) employing a Video-LMM as a direct judge by prompting it with our human annotation guidelines, and (2) computing multimodal embedding similarity between the original video and generated summary using a state-of-the-art multimodal encoder\footnote{We use LanguageBind~\citep{zhu2023languagebind} for multimodal embedding in this paper.}.

\paragraph{Results and Analysis.} Table~\ref{tab:correlation-results} presents correlations measured by Kendall's $\tau_b$, $\tau_c$, and Spearman's $\rho$. QEVA consistently achieves significantly higher correlations than all existing metrics, highlighting its superior alignment with human evaluations, especially notable given QEVA is a reference-free metric.

\subsection{Model-wise Correlation Analysis} \label{sec:modelwise-corr}

To further examine QEVA's consistency, we analyze correlation results separately for each Video-LMM summarization model. Table~\ref{tab:model-wise correlation} shows that QEVA consistently yields positive and stable correlations across different summarization models. In contrast, some baseline metrics even show negative or inconsistent correlations for certain models.

We observe relatively lower correlations for Video-LLaVA-generated summaries. Upon qualitative analysis, we find these summaries are generally shorter and of lower quality, making it challenging for QEVA to generate meaningful questions and answers.

\subsection{Ablation Studies} \label{sec:ablation}

We perform ablation studies to analyze QEVA's robustness and generalizability.

\paragraph{Evaluation Criteria-wise Ablation.} 
We separately calculate correlations between human judgments and QEVA scores for each evaluation criterion (Coverage, Factuality, Chronology). As shown in Table~\ref{tab:criterion-ablation}, QEVA demonstrates strong correlations across all individual criteria, validating its capability to accurately capture different aspects of summarization quality.

\input{attachments/table3}
\input{attachments/table4}

\paragraph{Robustness to Different Video-LMM and LLM Models.} 
We further examine QEVA's internal robustness by replacing the original Video-LMM (Gemini-1.5 Pro) and LLM (GPT-4o) components with alternative models. Specifically, we test open-source Video-LMM variants (QwenVL, InternVL) and LLM variants (LLaMA, Gemma). Table~\ref{tab:component-ablation} indicates that QEVA maintains high correlations even with open-source alternatives. This demonstrates QEVA's practical applicability and cost-effectiveness by alleviating reliance on costly API-based models.

%% file: attachments/table1.tex
\begin{table*}[h!]
    \centering
\resizebox{\textwidth}{!}{%
\begin{tabular}{llcllll}
\hline
\multirow{2}{*}{\textbf{Direction}} &
  \multirow{2}{*}{\textbf{Metric}} &
  \multirow{2}{*}{\textbf{\begin{tabular}[c]{@{}c@{}}Reference\\ Summary\end{tabular}}} &
  \multicolumn{1}{c}{\multirow{2}{*}{\textbf{\begin{tabular}[c]{@{}c@{}}Video\\ Used\end{tabular}}}} &
  \multicolumn{3}{c}{\textbf{MLVU(VS)-Eval}} \\ \cline{5-7} 
 &
   &
   &
  \multicolumn{1}{c}{} &
  \multicolumn{1}{c}{\textbf{Kendall's $\tau_b$}} &
  \multicolumn{1}{c}{\textbf{Kendall's $\tau_c$}} &
  \multicolumn{1}{c}{\textbf{Spearman's $\rho$}} \\ \hline
\multirow{5}{*}{Rule-based}       & BLEU-1                   & \checkmark            &                               & 0.0217 (0.7803) & 0.0219 (0.7803) & 0.0431 (0.7045) \\
                                  & BLEU-2                   & \checkmark            &                               & 0.0673 (0.3865) & 0.0680 (0.3865) & 0.1045 (0.3561) \\
                                  & BLEU-3                   & \checkmark            &                               & 0.1994 (0.0152) & 0.1836 (0.0152) & 0.2836 (0.0108) \\
                                  & BLEU-4                   & \checkmark            &                               & 0.2113 (0.0155) & 0.1578 (0.0155) & 0.2798 (0.0119) \\
                                  & ROUGE-L                  & \checkmark            &                               & 0.1094 (0.1593) & 0.1104 (0.1593) & 0.1571 (0.1639) \\ \hline
\multirow{5}{*}{Similarity-based} & METEOR                   & \checkmark            &                               & 0.2663 (0.0006) & 0.2689 (0.0006) & 0.3822 (0.0005) \\
                                  & CIDEr                    & \checkmark            &                               & 0.2401 (0.0015) & 0.2435 (0.0014) & 0.3612 (0.0012) \\
                                  & SPICE                    & \checkmark            &                               & 0.2287 (0.0021) & 0.2312 (0.0020) & 0.3489 (0.0019) \\
                                  & BERTscore                & \checkmark            &                               & 0.1415 (0.0688) & 0.1428 (0.0688) & 0.1987 (0.0773) \\
                                  & Video-Summary Similarity & \multicolumn{1}{l}{} & \multicolumn{1}{c}{\checkmark}                     & 0.0278 (0.0517) & 0.0266 (0.0522) & 0.0401 (0.0549) \\ \hline
\multirow{2}{*}{LLM-based}        & MLVU                     & \checkmark            &                               & 0.5284 (0.0000) & 0.5309 (0.0000) & 0.6738 (0.0000) \\
                                  & Video-LMM Judge          & \multicolumn{1}{l}{} & \multicolumn{1}{c}{\checkmark} & 0.5376 (0.0000) & 0.5441 (0.0000) & 0.6810 (0.0000) \\ \hline
QA-based                          & QEVA(Ours)               & \multicolumn{1}{l}{} & \multicolumn{1}{c}{\checkmark} & \textbf{0.6465 (0.0000)} & \textbf{0.6407 (0.0000)} & \textbf{0.7326 (0.0000)} \\ \hline
\end{tabular}%
}
\caption{Comparative evaluation results of QEVA and existing summarization evaluation metrics on the MLVU(VS)-Eval benchmark. We report correlations with human judgments using Kendall's $\tau_b$, $\tau_c$, and Spearman's $\rho$. The metrics are categorized into several groups: rule-based n-gram metrics (BLEU variants, ROUGE-L), similarity-based metrics (METEOR, CIDEr, SPICE, BERTScore, Video-Summary Similarity), LLM-based metrics (MLVU, Video-LMM Judge), and our proposed QA-based metric (QEVA). Reference usage and video modality usage for each metric are also indicated.
\textbf{Takeaway:} QEVA consistently achieves significantly higher correlation with human judgments compared to existing metrics, demonstrating its effectiveness as a reference-free and multimodal evaluation metric.}
    \label{tab:correlation-results}
    \end{table*}

%% file: attachments/table2.tex
% Please add the following required packages to your document preamble:
% \usepackage{multirow}
% \usepackage{graphicx}
\begin{table*}[h!]
\centering
\resizebox{\textwidth}{!}{%
\begin{tabular}{l|cccc|cccc}
\hline
\multirow{2}{*}{\textbf{Metric}} &
  \multicolumn{4}{c|}{\textbf{Kendalls' $\tau_b$}} &
  \multicolumn{4}{c}{\textbf{Kendalls' $\tau_c$}} \\ \cline{2-9} 
 &
  \textbf{Qwen2.5-VL-7B} &
  \textbf{InternVL3-8B} &
  \textbf{Video-LLaVA-7B} &
  \textbf{GPT-4o} &
  \textbf{Qwen2.5-VL-7B} &
  \textbf{InternVL3-8B} &
  \textbf{Video-LLaVA-7B} &
  \textbf{GPT-4o} \\ \hline
BLEU-1                   & 0.1662 & 0.0870  & 0.0758          & 0.0053 & 0.1669  & 0.0867  & 0.0758          & 0.0054 \\
BLEU-2                   & 0.2735 & 0.1247  & 0.1457          & 0.1333 & 0.2746  & 0.1246  & 0.1458          & 0.1339 \\
BLEU-3                   & 0.2914 & 0.1380  & 0.1370          & 0.3293 & 0.2800  & 0.1300  & 0.1108          & 0.3088 \\
BLEU-4                   & 0.1654 & 0.1130  & \textbf{0.3565} & 0.1717 & 0.1444  & 0.0914  & 0.2025          & 0.1463 \\
ROUGE-L                  & 0.4022 & 0.2332  & 0.1573          & 0.3423 & 0.4038  & 0.2329  & 0.1575          & 0.3429 \\ \hline
METEOR                   & 0.2628 & 0.1573  & 0.3555          & 0.4107 & 0.2638  & 0.1571  & \textbf{0.3558} & 0.4125 \\
BERTscore                & 0.5202 & 0.0163  & 0.0701          & 0.3147 & 0.5223  & 0.0163  & 0.0700          & 0.3161 \\
Video-Summary Similarity & 0.0012 & -0.0301 & 0.2277          & 0.2417 & -0.0218 & -0.0119 & 0.2117          & 0.1472 \\ \hline
MLVU                     & 0.3950 & 0.3260  & -0.1254         & 0.2361 & 0.3900  & 0.3208  & -0.1067         & 0.2392 \\
Video-LMM Judge          & 0.4138 & 0.3358  & 0.2056          & 0.2178 & 0.2005  & 0.3123  & 0.1233          & 0.2198 \\ \hline
\textbf{QEVA(Ours)} &
  \textbf{0.4509} &
  \textbf{0.4268} &
  0.1450 &
  \textbf{0.4262} &
  \textbf{0.4500} &
  \textbf{0.4114} &
  0.1000 &
  \textbf{0.4222} \\ \hline
\end{tabular}%
}
\caption{Model-wise correlation analysis of summarization evaluation metrics on the MLVU(VS)-Eval dataset. Correlations (measured by Kendall's $\tau_b$ and $\tau_c$) are reported separately for summaries generated by four representative Video-LMM models: Qwen2.5-VL-7B, InternVL3-8B, Video-LLaVA-7B, and GPT-4o.
\textbf{Takeaway:} QEVA exhibits stable and positive correlations with human judgments across diverse Video-LMM summarization models, highlighting its robustness and generalizability.}
\label{tab:model-wise correlation}
\end{table*}

%% file: attachments/table3.tex
\begin{table}[h!]
\centering
\begin{tabularx}{\columnwidth}{l|>{\centering\arraybackslash}X >{\centering\arraybackslash}X >{\centering\arraybackslash}X}
\hline
\textbf{Criteria} & \textbf{$\tau_b$} & \textbf{$\tau_c$} & \textbf{$\rho$} \\ 
\hline
\textbf{Coverage}   & 0.6005 & 0.5872 & 0.7349 \\
\textbf{Factuality} & 0.5731 & 0.5586 & 0.7123 \\
\textbf{Chronology} & 0.5610 & 0.5479 & 0.6984 \\
\hline
\end{tabularx}
\caption{Ablation study of QEVA across the three individual evaluation criteria (Coverage, Factuality, Chronology). Correlations with human judgments (measured by Kendall's $\tau_b$, $\tau_c$, and Spearman's $\rho$) are reported separately for each dimension. 
\textbf{Takeaway:} QEVA demonstrates strong and consistent correlations across all individual evaluation criteria, validating its comprehensive and multidimensional evaluation approach.}
\label{tab:criterion-ablation}
\end{table}

%% file: attachments/table4.tex
% Please add the following required packages to your document preamble:
% \usepackage{graphicx}
% \usepackage[normalem]{ulem}
% \useunder{\uline}{\ul}{}
\begin{table}[t]
\centering
\resizebox{\columnwidth}{!}{%
\begin{tabular}{l|ccc}
\hline
\textbf{Video-LMM + LLM}         & \textbf{$\tau_b$} & \textbf{$\tau_c$} & \textbf{$\rho$}  \\ \hline
Qwen2.5-VL + LLaMA-3.1           & 0.6211            & 0.6152            & 0.7012          \\
Qwen2.5-VL + Gemma-3             & 0.6075            & 0.6021            & 0.7120          \\
InternVL3 + LLaMA-3.1            & 0.5988            & 0.5933            & 0.6761          \\
InternVL3 + Gemma-3              & 0.5827            & 0.5780            & 0.6894          \\
\textbf{Gemini-1.5-Pro + GPT-4o} & \textbf{0.6465}   & \textbf{0.6407}   & \textbf{0.7326} \\ \hline
\end{tabular}%
}
\caption{Ablation study examining the robustness of QEVA to different combinations of Video-LMM and LLM models compared to the default setting (Gemini-1.5 Pro + GPT-4o). We report correlations with human annotations (Kendall's $\tau_b$, $\tau_c$, and Spearman's $\rho$) when replacing original models with alternative open-source models (Qwen2.5-VL, InternVL3, LLaMA-3.1, Gemma-3).
\textbf{Takeaway:} QEVA maintains high correlations with human judgments even when using alternative open-source models, indicating practical applicability and cost-effectiveness without relying solely on costly API-based models.}
\label{tab:component-ablation}
\end{table}

%% file: content/5_con.tex
\section{Conclusion}
\label{sec:conclusion}

We propose QEVA, a novel reference-free metric for evaluating narrative video summarization leveraging multimodal question answering. 
QEVA demonstrates significantly higher correlation with human judgments compared to existing metrics, while eliminating reliance on costly reference summaries. 
Our approach facilitates scalable, accurate, and practical evaluation of video summarization systems, accelerating the development and deployment of Video-LMMs in real-world multimodal applications.

%% file: content/6_lim.tex
\section*{Limitations}

Despite the effectiveness and practicality demonstrated by QEVA, our proposed metric inherits several limitations inherent to its underlying models and design.

\paragraph{Hallucination.} 
As QEVA employs a Large Multimodal Model (LMM) for question generation and answering, it may occasionally produce hallucinated content in the generated questions or answers not actually present in the video. Although our filtering process significantly mitigates this issue, the possibility of subtle hallucinations remains, potentially affecting evaluation reliability in edge cases.

\paragraph{API Cost and Processing Speed.}
QEVA relies heavily on inference from Large Language Models (LLMs) and Video-LMMs. Such models typically require significant computational resources and incur relatively high API costs, particularly when evaluating large-scale datasets or numerous summaries. This dependence may limit QEVA's practical applicability in resource-constrained environments or real-time scenarios.

\paragraph{Necessity of Post-processing.}
QEVA occasionally produces outputs that deviate from the predefined format or scoring criteria. Although infrequent, these cases necessitate additional post-processing to ensure compliance with the intended evaluation guidelines, slightly complicating the evaluation pipeline.

\paragraph{Preference for LLM-based Outputs.}
Recent evaluations using LLMs have identified a subtle preference bias towards outputs generated by LLMs themselves. QEVA may similarly exhibit a slight bias favoring summaries produced by certain Video-LMMs, potentially influencing the fairness and objectivity of the evaluation. This phenomenon warrants further investigation to quantify and mitigate such biases in future research.

\section*{Ethics Statement}
Ethics Statement
Our research involved a user study with human participants to collect judgments for evaluating our proposed metric. The study was conducted in accordance with established ethical guidelines. We recruited 20 annotators who participated on a voluntary basis. While our institution's review board determined that formal IRB approval was not required for this type of user study (involving subjective evaluation of anonymized system outputs), we took several measures to ensure the protection of participants.

Prior to the study, all participants were provided with a clear description of the research objectives and the annotation task, and we obtained informed consent from each individual. To acknowledge their valuable contribution, participants were fairly compensated for their time and effort. All data collected during the study was fully anonymized to protect the privacy and confidentiality of the participants, and no personally identifiable information (PII) was collected or stored.

\section*{Acknowledgement}
This work was supported by IITP grant funded by the Korea government(MSIT) (No. RS-2020-II200004, Development of Previsional Intelligence based on Long-term Visual Memory Network) and supported by the Institute of Information \& Communications Technology Planning \& Evaluation (IITP) grand funded by the Korea government (MSIT) [RS-2021-II211341, Artificial Intelligence Graduate School Program(Chung-Ang University)].

%% file: content/appendix.tex
\appendix

\section{Full Prompts}
\label{sec:full-prompts}
In this section, we provide the complete prompts utilized for generating QA pairs across the three distinct QEVA evaluation dimensions—\textbf{Coverage}, \textbf{Factuality}, and \textbf{Chronology}. We present these prompts exactly as they were given to the Large Multimodal Models (Video-LMMs) and Large Language Models (LLMs) in our experiments. Researchers can use these prompts to precisely reproduce the QA generation process described in Section~\label{section:3.2} of the main paper.

\input{attachments/appendix/appendix-a1}
\input{attachments/appendix/appendix-a2}
\input{attachments/appendix/appendix-a3}

%% file: attachments/appendix/appendix-a1.tex
\begin{figure*}[t]
    \centering
    \includegraphics[width=\textwidth]{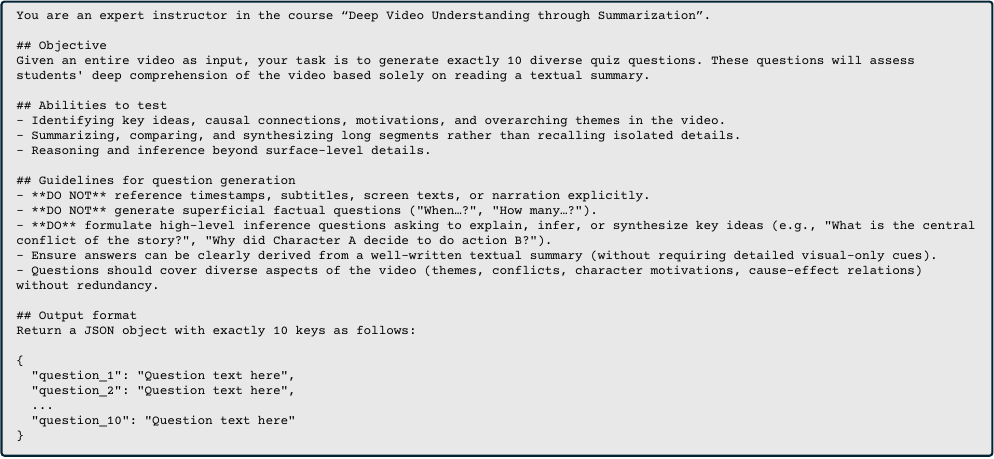}
    \caption{Methodology figure of QEVA.}
    \label{fig:qeva_pipeline}
    \end{figure*}

%% file: attachments/appendix/appendix-a2.tex
\begin{figure*}[t]
    \centering
    \includegraphics[width=\textwidth]{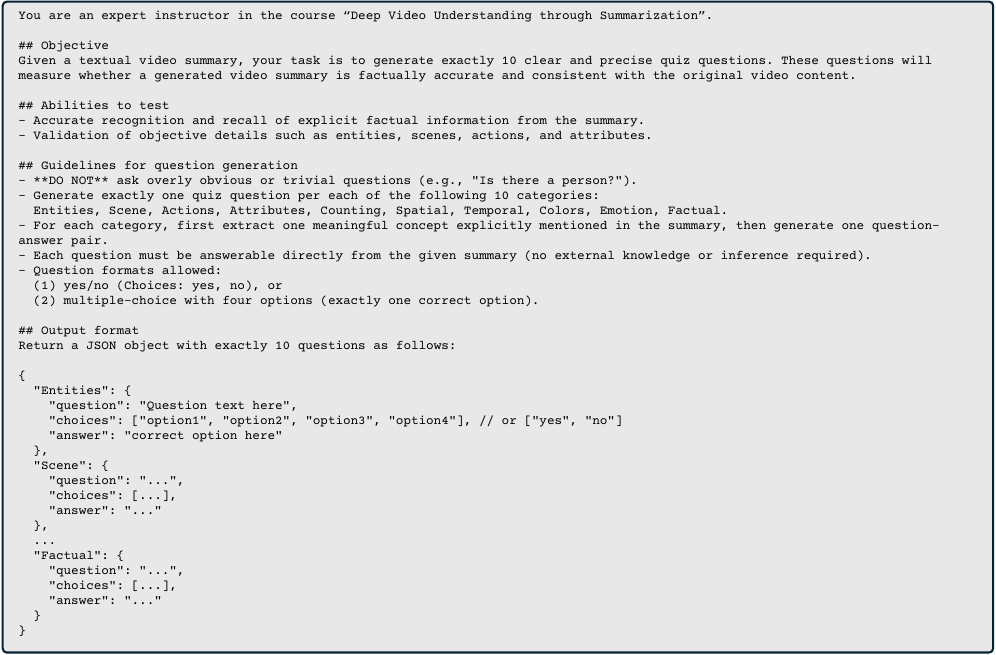}
    \caption{Methodology figure of QEVA.}
    \label{fig:qeva_pipeline}
    \end{figure*}

%% file: attachments/appendix/appendix-a3.tex
\begin{figure*}[t]
    \centering
    \includegraphics[width=\textwidth]{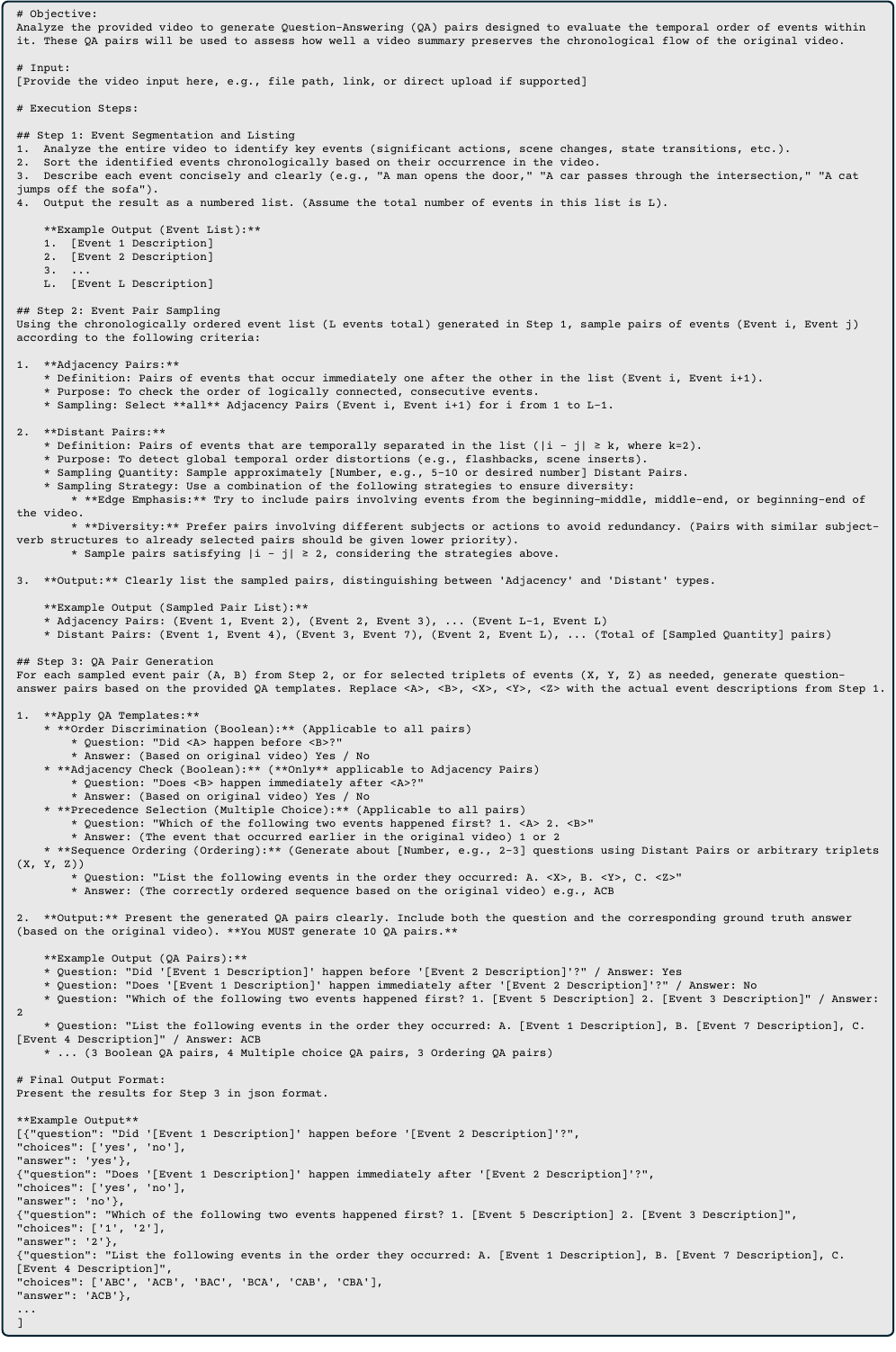}
    \caption{Methodology figure of QEVA.}
    \label{fig:qeva_pipeline}
    \end{figure*}